\documentclass[11pt,a4paper]{article}
\usepackage[hyperref]{eacl2021}
\usepackage{times}
\usepackage{latexsym}

\usepackage{booktabs}
\usepackage{multicol}
\usepackage{makecell}
\usepackage{amsmath}
\usepackage{amssymb}
\usepackage{latexsym}
\usepackage{xspace}

\usepackage{graphicx}
\usepackage{tabularx}
\usepackage{url}
\usepackage{multirow}
\usepackage{float}
\usepackage[inline]{enumitem}
\usepackage{pgffor}

\usepackage{xcolor}

\usepackage{microtype}

\aclfinalcopy 

\newif{\ifhidecomments}
\hidecommentsfalse

\ifhidecomments
    \newcommand{\akos}[1]{}
    \newcommand{\judit}[1]{}
    \newcommand{\andras}[1]{}
\else
    \newcommand{\akos}[1]{\textcolor{pink}{[#1 ({\bf \'Akos})]}}
    \newcommand{\judit}[1]{\textcolor{blue}{[#1 ({\bf Judit})]}} 
    \newcommand{\andras}[1]{\textcolor{orange}{[#1 ({\bf Andr\'as})]}} 
\fi
\newif{\ifhidetodo}
\hidetodofalse

\ifhidetodo
    \newcommand{\todo}[1]{}
\else
    \newcommand{\todo}[1]{\textcolor{magenta}{[TODO: #1 ]}}
\fi
\newcommand{\com}[1]{}
\newcommand{\camready}[1]{}
\newcommand{\resolved}[1]{} 
\newcommand{\task}[3]{{$\langle$#1, #2, #3$\rangle$\xspace}}

\newcommand{\mbert}{mBERT\xspace}
\newcommand{\mroberta}{XLM-RoBERTa\xspace}
\newcommand{\wordpiece}{wordpiece\xspace}

\newcommand{\swchoice}[1]{\textsc{#1}\xspace}
\newcommand{\swlstm}{\swchoice{lstm}}
\newcommand{\swmlp}{\swchoice{attn}}

\usepackage{microtype}

\title{Subword Pooling Makes a Difference}

\author{Judit \'Acs$^{1,3}$\\
    Budapest University of\\ Technology and Economics$^1$\\
    \texttt{judit@aut.bme.hu} \\\And
    \'{A}kos K\'{a}d\'{a}r$^2$ \\
    Borealis AI$^2$ \\
    \texttt{akos.kadar@}\\ \texttt{borealisai.com}\\\And
    Andr\'as Kornai$^{1,3}$\\
    Institute for Computer Science\\ and Control$^3$ \\
    \texttt{andras@kornai.com}}

\date{}

\begin{document}
\maketitle

\begin{abstract}

Contextual word-representations became a standard in modern natural language
processing systems. These models use subword tokenization to handle large
vocabularies and unknown words. Word-level usage of such systems requires a
way of pooling multiple subwords that correspond to a single word. 
In this paper we investigate how the choice of subword pooling affects the
downstream performance on three tasks: morphological probing, POS tagging
and NER, in 9 typologically diverse languages. We compare these in two massively
multilingual models, \mbert and \mroberta
. For morphological tasks, the widely
used `choose the first subword' is the worst strategy and the best results are
obtained by using attention over the subwords. For POS tagging both of
these strategies perform poorly and the best choice is to use a small LSTM over
the subwords. The same strategy works best for NER and we show 
that \mbert is better than \mroberta in all 9 languages. We publicly release
all code, data and the full result tables at
\url{https://github.com/juditacs/subword-choice}.
\end{abstract}

\section{Introduction}

Training of contextual language models on large training corpora
generally begins with segmenting the input into subwords \cite{Schuster:2012} to reduce the vocabulary size. Since most tasks
consume full words, practitioners have the freedom to decide whether to use
the first, the last, or some combination of all subwords. The original paper
introducing BERT, \citet{Devlin:2018a}, suggests using the first subword for
named entity recognition (NER), and did not explore different poolings.
\citet{Kondratyuk:2019} also use the first subword, for dependency parsing,
and remark in a footnote that they tried the first, last, average, and max
pooling but the choice made no difference. \citet{Kitaev:2019} report similar
findings for constituency parsing, but nevertheless opt for reporting results only
using the last subword. \citet{Hewitt:2019} take the average of the subword
vectors for syntactic and word sense disambiguation tasks. \citet{Wu:2020} use
attentive pooling with a trainable norm for news topic classification and
sentiment analysis in English. \citet{Shen:2018b} use hierarchical pooling for
sequence classification tasks in English and Chinese.

Here we show that for word-level tasks (morphological, POS and NER tagging),
particularly for languages where the proportion of multi-subword tokens
(i.e. those word tokens that are split into more than one subword) is high,
more care needs to be taken as both pooling strategy, and that the choice of
language matters. We demonstrate this clearly for European languages with rich
morphology, and in Chinese, Japanese and Korean (CJK).  Similar to subword
pooling, the choice of the lowest layer, the topmost one, or some combination
of the activations in different layers has to be made. Here our main focus is
subword pooling, but we do discuss layer pooling to the extent it sheds
light on our main topic. We observe that the gap between using the first
and the last subword unit is larger in lower layers than in higher ones.


We describe our data and tasks in Section~\ref{sec:data}, and the 
subword pooling strategies investigated in Section~\ref{sec:subword_pooling}.
Our results are presented in Section~\ref{sec:results}, and in
Section~\ref{sec:conclusion} we offer our conclusions. 

Our main contributions are:

\begin{itemize}
    \item we show that subword pooling matters, the differences between choices
        are often significant and not always predictable;
    \item \mroberta \cite{Conneau:2019a} is slightly better than \mbert in the majority of
        morphological and POS tagging tasks; while \mbert is better at NER in
        all languages;
    \item the common choice of using the first subword is generally worse than
        using the last one for morphology and POS but the best for NER;
    \item the difference between using the first and the last subword is larger
        in lower layers than in higher layers and it is more pronounced in
        languages with rich morphology than in English;
    \item the choice of subword pooling makes a large difference for
        morphological and POS tagging but it is less important for NER;
    \item we release the code, the data and the full result
        tables.
\end{itemize}

\section{Tasks, languages, and architectures}\label{sec:data}

We investigate pooling through three kinds of tasks. In {\it morphological}
tasks we attempt to predict morphological features such as gender, tense, or
case. In {\it POS} tasks we predict the lexical category associated with each
word. In {\it NER} tasks we assign BIO tags \cite{Ramshaw:1995} to named entities. We chose
word-level, as opposed to syntactic, tasks because they can be tackled with
fairly simple architectures and thus allow for a large number of experiments
that highlight the differences between subword pooling strategies. Our
experiments are limited only by the availability of standardized multilingual
data.

We use Universal Dependencies (UD) \cite{Nivre:2018s} for morphological and POS tasks, and WikiAnn
\cite{Pan:2017} for NER.  We pick the largest treebank in each language from UD
and sample 2000 train, 200 dev and 200 test sentences for the morphological
probes and up to 10,000 train, 2000 dev and 2000 test sentences -- often
limited by the size of the treebank -- for POS.  We chose languages with
reasonably large treebanks in order to generate enough training data, making
sure we have an example from each language family, as well as one from European
subfamilies since their treebanks tend to be very large.  We use 10,000 train,
2000 dev and 2000 test sentences for NER.  Preprocessing steps are further
explained in Appendix~\ref{app:data}.  Our choice of languages are Arabic,
Chinese, Czech, English, Finnish, French, German, Japanese, and Korean. UD's
gold tokenization is kept and we run subword tokenization on individual tokens
rather then the full sentences.

\paragraph{Morphological tasks} UD assigns zero or more tag-value pairs to
each token such as \verb|VerbForm=Ger| for `asking'.  We define a probe as a
triplet of \task{language}{tag}{POS}, i.e.~we train a classifier to predict the
value of a single tag in a sentence in a particular
language.\footnote{Consolidating these triplets across POS would be misleading
in that the results show large variation across different POS values.}  The task
\task{English}{VerbForm}{VERB} would be trained to predict one of three labels
for each English verb: finite, infinite or gerund.  We pick 4 tasks that are
applicable to at least 3 of the 6 languages where the task makes sense (there
are no morphological tags for Chinese and Japanese, and Korean uses a different
tagging scheme). 
Table~\ref{tab:morphology_tasks} lists the probing tasks.

\paragraph{Part-of-speech tagging} assigns a syntactic category to each token
in the sentence. Usually treated as a crucial low level task to provide useful
features for higher level linguistic analysis such as syntactic and semantic
parsing.  Universal POS tags (UPOS) are available in UD in all 9 languages.  

\paragraph{Named entity recognition} is a classic information extraction
subtask that seeks to identify the span of named entities mentioned in the
sentence and classify them into pre-defined categories such as person names,
organizations, locations etc. NER was the only token level task explored in
the original BERT paper \citet{Devlin:2018a}.

\paragraph{Architectures} BERT and other contextual models use subword
tokenizers that generate one or more subwords for each token. In this study we
compared \mbert and \mroberta, two Transformer-based large scale language
models with support for over 100 languages. We pick these two since they are
architecturally similar (both have 12 layers and the same hidden size) making
our comparison easier.  \mbert was trained on Wikipedia while \mroberta was
trained on CommonCrawl \cite{Wenzek:2020}.  Both models have been extensively
applied to English and multilingual tasks, but generally at the sentence or
sentence pair level, where subword issues do not come to the fore. \mbert uses
a common \wordpiece vocabulary with 118k subword units.  When a word is split
into multiple subword units, each token that is not the first one is prefixed
with \verb|##|.  \mroberta's vocabulary was trained in a similar fashion but
with 250k units and a special start symbol (Unicode lower eights block) instead
of continuation symbols.  Each word is prefixed with this start symbol before
it is tokenized into one or more subword units.  These start symbols are often
then tokenized as single units, particularly before Chinese, Japanese and
Korean characters, therefore artificially increasing the subword unit count.
We indicate the proportion of words starting with a standalone start symbol
along with other tokenization statistics in Table~\ref{tab:wp_stats}.

\begin{table}
    \centering
\begin{tabular}{lllr}
\toprule
{\bf Language} &       {\bf Tag} &   {\bf POS} &  {\bf \# class} \\
\midrule
  Arabic &      case &  NOUN &        3 \\
  Arabic &    gender &   ADJ &        2 \\
\midrule
   Czech &    gender &   ADJ &        3 \\
   Czech &    gender &  NOUN &        3 \\
\midrule
 English &  verbform &  VERB &        4 \\
\midrule
 Finnish &      case &  NOUN &       12 \\
 Finnish &  verbform &  VERB &        3 \\
\midrule
  French &    gender &   ADJ &        2 \\
  French &    gender &  NOUN &        2 \\
  French &  verbform &  VERB &        3 \\
\midrule
  German &      case &  NOUN &        4 \\
  German &    gender &   ADJ &        3 \\
  German &    gender &  NOUN &        3 \\
  German &  verbform &  VERB &        3 \\
\bottomrule
\end{tabular}
\caption{\label{tab:morphology_tasks} List of morphological probing tasks. The last column is the number of classes in a particular task.}
\end{table}

\begin{table}
\begin{tabular}{lrr|rrr}
\toprule
{} & \multicolumn{2}{c}{\bf {\mbert}} & \multicolumn{3}{c}{\bf {\mroberta}}\\
{} & count & 2+ & count & 2+ & \_start\\
\midrule

Arabic   &           1.95 &     48.9 &             1.49 &       35.0 &        3.4 \\
Chinese  &           1.58 &     53.5 &             2.13 &       88.5 &       86.6 \\
Czech    &           2.04 &     53.0 &              1.7 &       45.2 &        1.6 \\
English  &           1.25 &     14.3 &             1.25 &       16.9 &        0.8 \\
Finnish  &           2.32 &     67.3 &             1.86 &       53.0 &        2.3 \\
French   &           1.34 &     22.4 &             1.41 &       28.7 &        2.1 \\
German   &           1.64 &     30.6 &             1.57 &       29.7 &        1.3 \\
Japanese &            1.6 &     43.0 &             2.25 &       94.6 &       92.9 \\
Korean   &           2.44 &     75.7 &             2.16 &       67.3 &        9.0 \\

\bottomrule
\end{tabular}

\caption{\label{tab:wp_stats} Subword tokenization statistics by language
  and model. First and third columns: average number of pieces that one word
  is split into. Second and fourth columns: proportion of multi-subword
  words. Last column: proportion of words that start with a standalone start
  token in \mroberta.}

\end{table}

As Table~\ref{tab:wp_stats} shows, the number of subword tokens is highly
dependent on the language.  English words are only split in 14.3\%
(resp. 16.9\%) of the time by the two models, while in many other languages
more than half of the words are tokenized into two or more subword units.  We
hypothesize that this is due to the combination of the characteristics of the
English language and its overrepresentation in the training data and the
subword vocabulary.  

We also observe that the two models' tokenizers
work in very different ways. Out of the 2800 morphological test examples, only
58 are tokenized the same way and 51 of these are not split into multiple
subwords. Only 7 words that are in fact tokenized, are tokenized the same
way. Although the full tokenization is rarely the same, the first and the last
subwords are the same in 45.5\% and in 44.7\% of the cases.

\section{Subword pooling}\label{sec:subword_pooling}

We test 9 types of pooling methods listed in Table~\ref{tab:pooling_methods}
and grouped in three broad types.  The first group uses the first and last
subword representations in some combination.  In \swchoice{f+l} pooling the
mixing weight is the only learned parameter.  The second group are
parameter-free elementwise pooling operations.

\begin{table}[ht]
    \centering
        \begin{tabular}{lll}
            \toprule
             {\bf Method} & {\bf Explanation} & {\bf \hspace*{-5mm}Params} \\
             \midrule
             \swchoice{first} & first subword unit & none \\
             \swchoice{last} & last subword unit & none \\
             \swchoice{last2} & \makecell[l]{concatenation of the\\last two subword units} & none \\
             \swchoice{f+l} & $w u_\text{first} + (1-w) u_\text{last}$ & \hspace*{4mm}$w$ \\
            \midrule
            \swchoice{sum} & elementwise sum & none \\
            \swchoice{max} & elementwise max & none \\
            \swchoice{avg} & elementwise average & none \\
            \midrule
            \swchoice{attn} &  \makecell[l]{Attention over the\\ subwords, weights\\ generated by an MLP} & MLP \\
            \swchoice{lstm} & \makecell[l]{biLSTM reads all vectors,\\final hidden state}
            & \hspace*{-3mm}LSTM \\
            \bottomrule
        \end{tabular}
    \caption{Subword unit pooling methods. $u_\text{first}$ and $u_\text{last}$
    refer to the first and the last units respectively.}
    \label{tab:pooling_methods}
\end{table}

The last two methods rely on small neural networks that learn to combine the
subword representations.  Our subword {\sc attn} has one hidden layer of 50
neurons with ReLU activation and a final softmax layer that generates a
probability distribution over the subword units of the token. Similarly to
self-attention, these probabilities are used to compute the weighted sum of
subword representations to produce the final token vector.  The
\swchoice{lstm} uses a bi\-LSTM \cite{Hochreiter:1997} that summarizes the 768-dimensional vectors
(the hidden size of both models) into a 50-dimensional hidden vector in each
direction, which are then concatenated and passed onto the classifier. These
two are considerably more complicated and slower to train than the other
methods, but {\sc attn} works well for morphological tasks, and {\sc lstm} for
POS tagging in CJK languages. \citet{Shen:2018b} found hierarchical pooling
beneficial, but they investigated sentence level tasks where the subword
stream is much longer than in the word-level tasks we are considering (words
are rarely split into more than 4 subwords) and hierarchical pooling has
better traction. 

\begin{figure}[ht]
  \begin{center}
      \includegraphics[width=0.49\textwidth,trim={1cm 1cm 0.5cm 0.2cm}]{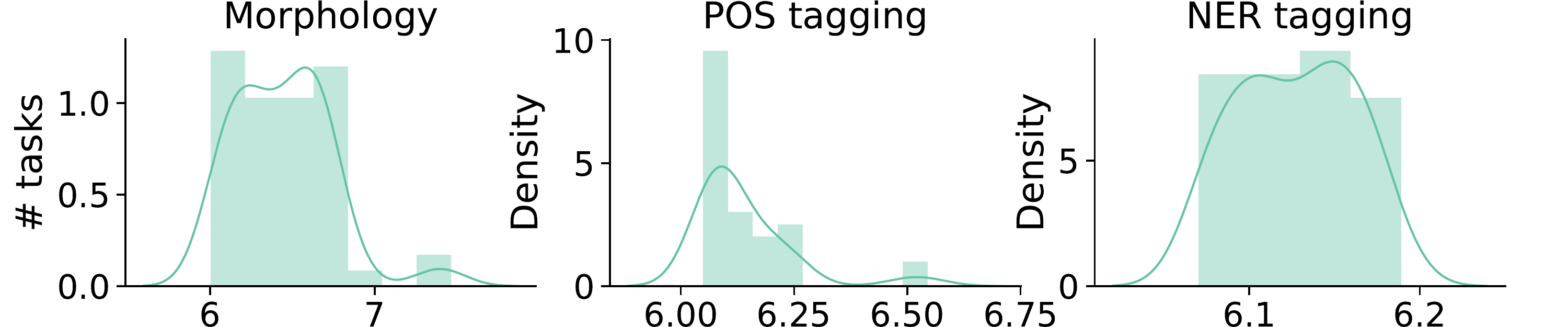}
  \end{center}
  \caption{Distribution of the weighted average of layers across all tasks.}
  \label{fig:dist_layer_average}
\end{figure}

\paragraph{Layer pooling effects} Both \mbert and \mroberta have an embedding layer
followed by 12 hidden layers.  The only contextual information available in
the embedding layer is the position of the token in the sentence. Hidden
activations are computed with the self-attention layers, therefore in theory
have access to the full sentence.  We ran our experiments for each
layer separately as well as for the sum of all layers.  For all tasks, as we move
up the layers, results also move up or down in tandem. As exhaustive
experiments considering different combinations of layers were computationally
too expensive for our setup, and would significantly complicate presentation
of our results, we pick a single setting for all experiments by computing the
best \emph{expected layer} for each task as

\begin{equation}
    \mathbb{E}(L) = \frac{\sum_{l_i \in L} i A(l_i)}{\sum_{l_i \in L} A(l_i)},
\end{equation}

\noindent
where $L$ is the set of all layers, $l_i$ is the $i$th layer, and $A(l_i)$ is
the development accuracy at layer $i$.  As Figure~\ref{fig:dist_layer_average}
shows, the expected layers are almost always centered around the 6th layer.
Therefore, with the exception of comparing {\sc first} and {\sc last}, which
we analyze in greater detail in \ref{ss:laypool}, we chose the 6th layer to
simplify the presentation.

\paragraph{Probing setup}

Every experiment is trained separately, with no parameter sharing between the
tasks and the experiments.  We probe the morphology on fixed representations
with a small MLP (multilayer perceptron) with a single hidden layer of 50 neurons and ReLU activation.
We train the same model for POS tagging and NER on top of each token representation.
We keep the number of parameters intentionally low, about 40k, to avoid
overfitting on the probing data and to force the MLP to probe the
representation instead of memorizing the data. We do note, however, that
\swmlp and \swlstm increase the number of trained parameters to 77k and 330k
respectively. We run each configuration 3 times with different random seeds.
The standard deviation of results is always less than 0.06 for morphology and
less than 0.005 for POS and NER.  Further details are available in Appendix~\ref{app:training}.

\paragraph{Choosing the size of the LSTM}

\swlstm is our subword pooling method with the most parameters. The number of
parameters scales quadratically with the hidden dimension of the LSTM. We pick
this dimension with binary parameter search on morphology tasks. Our early
experiments showed that increasing the size over 1000 showed no significant
improvement, and a binary search between 2 and 1024 led us to choose a 
biLSTM with 100 hidden units. 

\section{Results}\label{sec:results}

Our analysis consisted of two steps. We first performed the \swchoice{first}
and \swchoice{last} tasks at each layer (see
Figure~\ref{fig:line_morph_last_first_ratio}). Based on the results of this, 
we picked a single layer, the 6th, to test all 9 subword pooling choices. 
The full list of results on the 6th layer is listed in Appendix~\ref{app:full_results}.

\subsection{Layer pooling}\label{ss:laypool}

\begin{figure*}[ht]
  \begin{center}
      \includegraphics[width=0.98\textwidth,trim={2cm .5cm 0 0}]{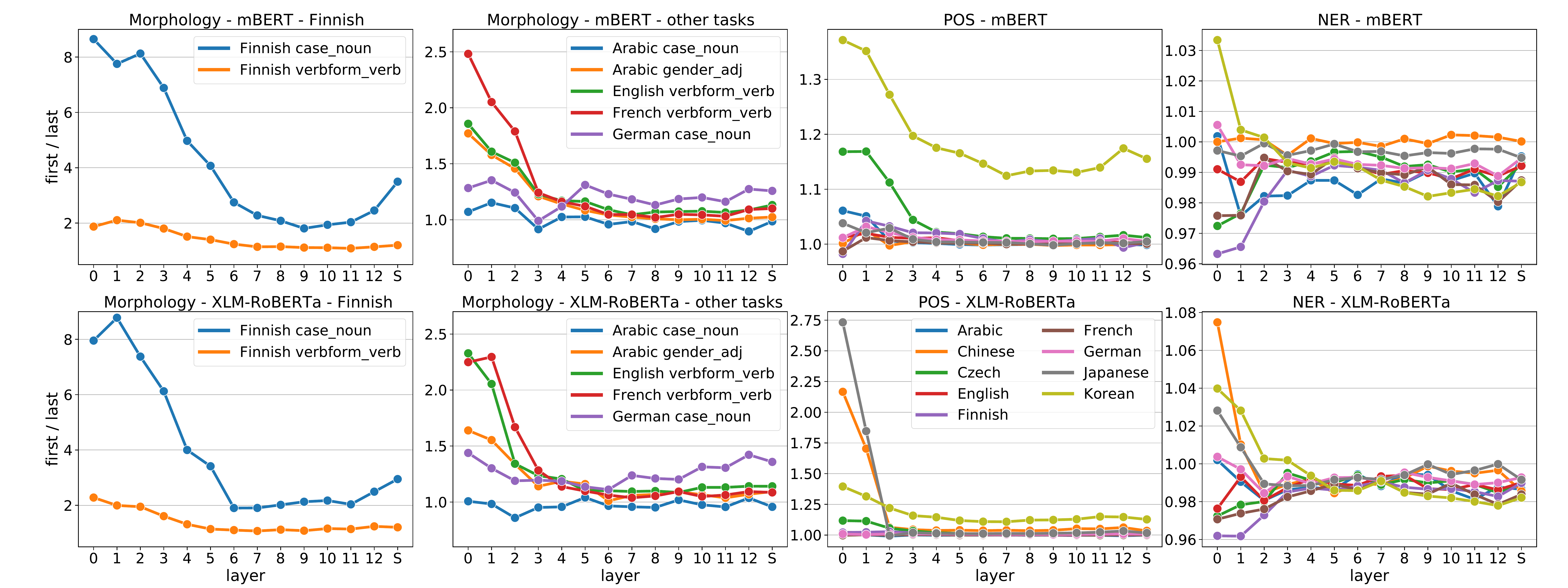}
  \end{center}
    \caption{\swchoice{last}-\swchoice{first} ratio of the test accuracy of
      some morphological tasks and of POS and NER in all languages across all
      layers. We plot Finnish morphological tasks separately since the
      effect is so pronounced that presenting them on the same plot would
      render the scaling uninformative for the other cases. S is the sum of
  all layers. Note that we do not have a strongly prefixing language due to
  the lack of available probing data.} 
  \label{fig:line_morph_last_first_ratio}
\end{figure*}

We find that although \swchoice{last} is almost always better than
\swchoice{first}, the gap is smaller in higher layers. We quantify this with
the ratio of the accuracy of \swchoice{last} and \swchoice{first} at the same
layer. Figure~\ref{fig:line_morph_last_first_ratio} illustrates this ratio for
a few selected morphological tasks and POS and NER for all 9 languages. We
split the morphological tasks into two groups, Finnish tasks and other tasks.
\task{Finnish}{Case}{NOUN} shows the largest gap in the lower layers,
\swchoice{last} is 8 times better than \swchoice{first}.  We observe smaller
gaps in other tasks. POS shows a fairly uniform picture with the exception of
Korean, where \swchoice{first} is worse in all layers and both models. Lower
layers in \mbert show a larger gap in Czech and the same is true for Chinese
and Japanese in \mroberta. NER shows little difference between \swchoice{first}
and \swchoice{last} except for the first few layers, particularly in Chinese
and Korean.  To interpret these results, keep in mind that CJK tokenization is
handled somewhat arbitrarily by \mroberta, particularly in the first subword
(c.f.~Table~\ref{tab:wp_stats}).


\subsection{Morphology}

We present the results of 14 morphological probing tasks (see
Table~\ref{tab:morphology_tasks}) and 9 subword pooling strategies (see
Table~\ref{tab:pooling_methods}) using the 6th layer of each model.

\begin{figure}[H]
  \begin{center}
  \includegraphics[width=0.45\textwidth,trim={0.5cm 0.7cm 0.5cm 0.cm}]{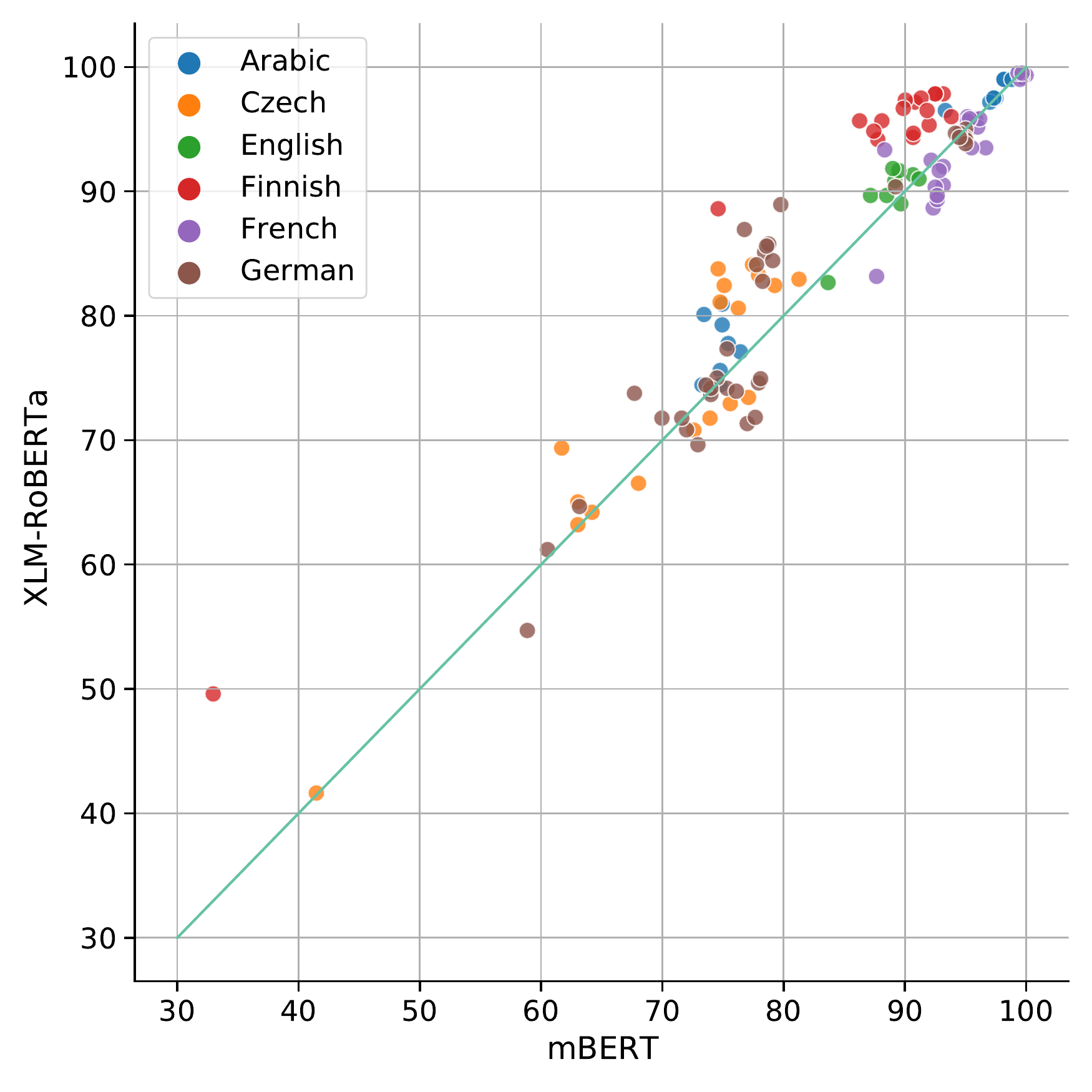}
  \end{center}
  \caption{Accuracy of \mbert vs.~\mroberta on morphological tasks.}
  \label{fig:scatter_morph_by_model}
\end{figure}

\paragraph{\mbert vs.~\mroberta} Averaging over all tasks, \mroberta
achieves 85.7\% macro accuracy while \mbert achieves 83.9\%.  On a
per-language basis, \mroberta is slightly better than \mbert except for
French.  Figure~\ref{fig:scatter_morph_by_model} shows our findings. The
two models generally perform similarly with the exception of French and
Finnish: \mbert is almost always better at French tasks, while \mroberta is
always better at Finnish tasks.  Similar trends emerge when looking at the
results by subword pooling method.  \mroberta is always better regardless of
the pooling choice but the difference is only significant ($p<0.05$) for
\swchoice{max} and \swchoice{sum}.\footnote{We use paired $t$-tests on the
  accuracy of the models on the 14 tasks.} These findings suggest that
\mroberta retains more about the orthographic presentation of a token, and
it uses tokenization that is closer to morpheme segmentation, hence performing
better at inflectional morphology, which is most often derivable from the word
form alone.

\begin{figure}[H]
  \begin{center}
      \includegraphics[width=.45\textwidth,trim={0.5cm 1.4cm 0.5cm 0.2cm}]{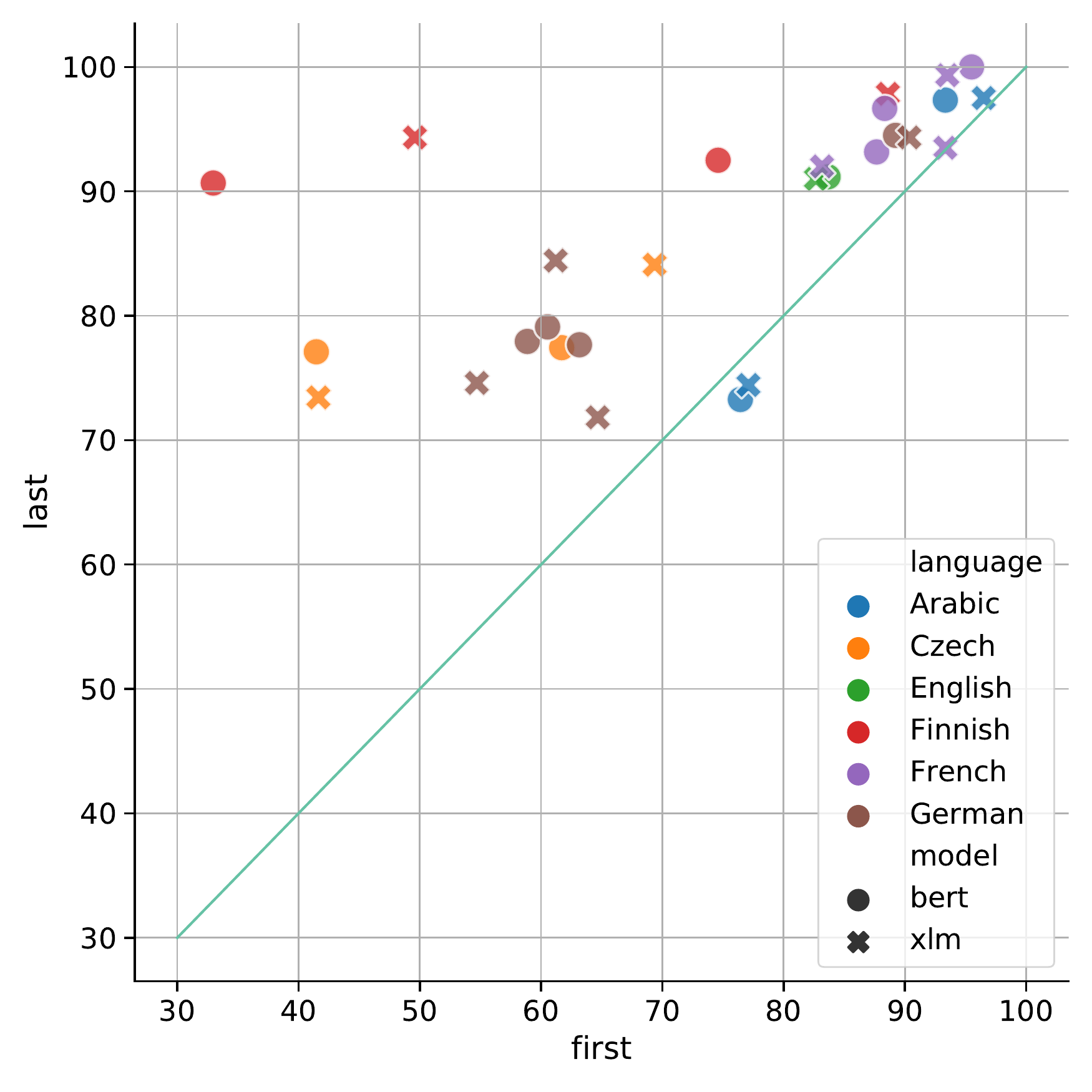}
  \end{center}
    \caption{\swchoice{first} vs.~\swchoice{last} on morphological tasks.}
  \label{fig:morph_first_last}
\end{figure}

\paragraph{First or last subword?} As Figure~\ref{fig:morph_first_last}
shows, with the exception of the \task{Arabic}{Case}{N} task, \swchoice{last}
is always better than \swchoice{first}.  We find the largest difference in
favor of \swchoice{last} in Finnish and Czech.
Table~\ref{tab:morph_first_last_extreme} lists all tasks where the difference
between \swchoice{first} and \swchoice{last} is larger than 20\% along with
the only counterexample (where the difference is about 10\% in the other
direction).  These findings are likely due to the fact that Finnish and Czech
exhibit the richest inflectional morphology in our sample.

The exceptional behavior of Arabic case may relate to the fact that case often
disappears in modern Arabic \cite{Biadsy:2009}. When this occurs the first
token, being closest to the previous word, may provide a more reliable
indicator, especially if that word was a preposition. Given the complex
distribution of Arabic case endings, our sample is too small to ascertain
this, and the results, about 75\% on a 3-way classification task, are
clearly too far from the optimum to draw any major conclusion (note that on
Finnish case, a 12-way classification task, we get above 94\%\footnote{Finnish has more than 12 cases but infrequent ones were excluded.}). 

\begin{figure*}[ht]
  \begin{center}
      \includegraphics[width=.99\textwidth,trim={0cm 0.8cm 0cm 0.3cm}]{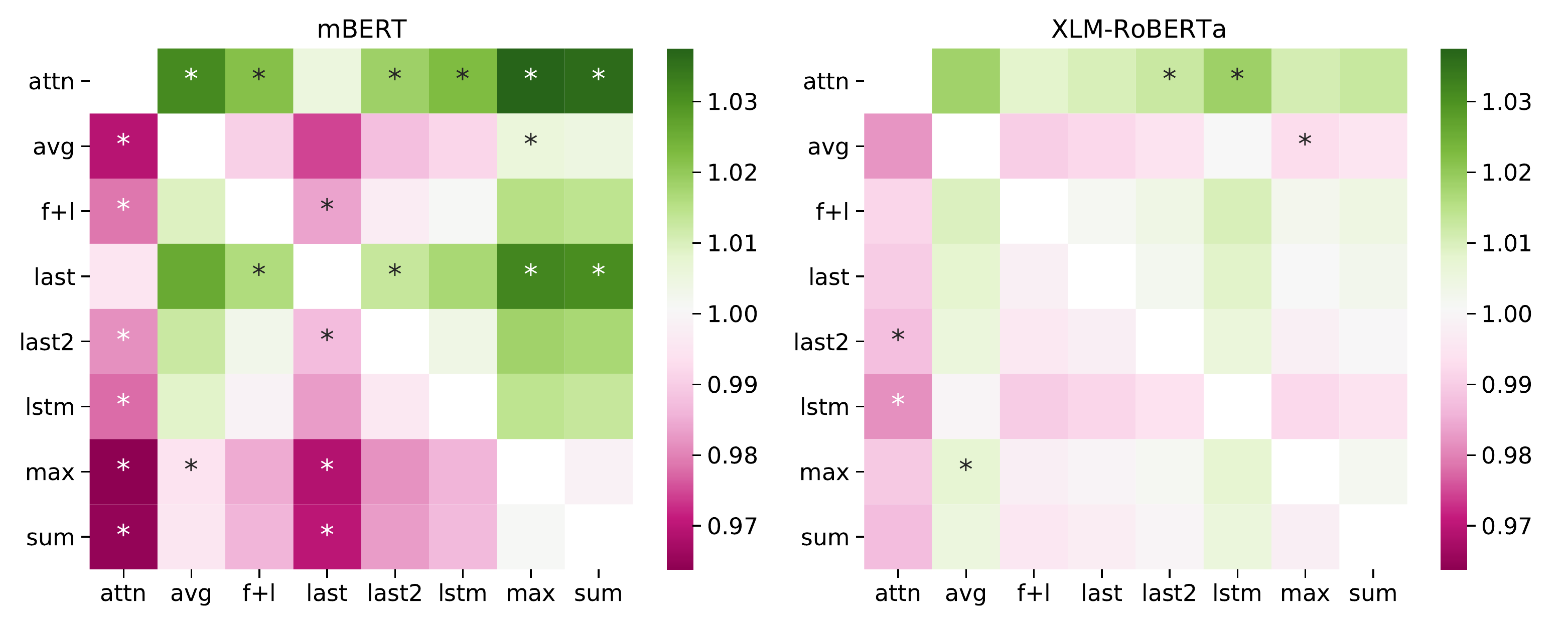}
  \end{center}
  \caption{Pairwise ratio of test accuracy by subword choice on morphological
    tasks. Colors indicate row/column ratios. Green cells mean that the row
    subword choice yields better results than the column choice. $\ast$ marks
    pairs where the difference is statistically significant. \swchoice{attn} is
    better than all other choices, therefore its row is
    green. \swchoice{first} is omitted for clarity as it is much worse than
    the other choices.}
  \label{fig:heatmap_pair_morph}
\end{figure*}

\paragraph{Other pooling choices}
While \swchoice{first} is clearly inferior in morphology, the picture is less
clear for the other 8 pooling strategies. As
Figure~\ref{fig:heatmap_pair_morph} illustrates, \swchoice{attn} is better than
all other choices for both models but its advantage is only significant over a
few other choices. We observe larger -- and more often significant --
differences in the case of \mbert than in \mroberta.

\begin{table}[h]
  \centering
    \footnotesize
\begin{tabular}{llrr}
\toprule
 task &    model &  first &  last \\
\midrule
    \task{Finnish}{Case}{N} &     \mbert &   33.0 &  90.7 \\
    \task{Finnish}{Case}{N} &  \mroberta &   49.6 &  94.4 \\
    \task{Czech}{Gender}{A} &   \mbert &   41.5 &  77.1 \\
    \task{Czech}{Gender}{A} &  \mroberta &   41.6 &  73.5 \\
    \task{German}{Gender}{N} &  \mroberta &   61.2 &  84.4 \\
    \midrule
    \task{Arabic}{Case}{N} & \mroberta &   77.1 &  74.5 \\
    \task{Arabic}{Case}{N} &  \mbert &   76.5 &  73.3 \\
\bottomrule
\end{tabular}
    \caption{Morphological tasks with the largest difference between \swchoice{first} and \swchoice{last}. The two tasks where \swchoice{first} is better than \swchoice{last} are also listed.
    \label{tab:morph_first_last_extreme}}
\end{table}

\paragraph{Attention weights}
The MLP used in \swchoice{attn} assigns a weight to each subword which are then normalized by
softmax. We examine these weights for each token in the test data for
morphology.
Table~\ref{tab:mlp_weights} lists the proportion of tokens where
\swchoice{attn} assigns the highest weight to the first, last or a middle token,
or the token is not split by the tokenizer. The last subword is weighted
highest in more than 80\% of the cases. The only task where the last subword
is not the most frequent winner is \task{Arabic}{Case}{N}, where the first is
weighted highest in 60\% of the tokens by both models. These findings are in
line with the behavior of \swchoice{first} and \swchoice{last}.

\begin{table}
  \centering
  \begin{tabular}{lrr}
\toprule
{} & XLM-RoBERTa &  mBERT \\
\midrule
first  &        7.1\% &   6.0\% \\
last   &       81.5\% &  83.7\% \\
middle &        5.9\% &   6.3\% \\
single &        5.5\% &   4.0\% \\
\bottomrule
\end{tabular}

  \caption{Distribution of the location of the highest weighted subword.
  Single refers to tokens that are not split by the tokenizer.}
  \label{tab:mlp_weights}
\end{table}

\subsection{POS tagging}

We train POS tagging models for 9 languages with 9 subword pooling strategies.
We evaluate the models using tag accuracy.

\begin{figure}[ht]
  \begin{center}
  \includegraphics[width=0.51\textwidth,trim={1cm 1cm .2cm 2cm}]{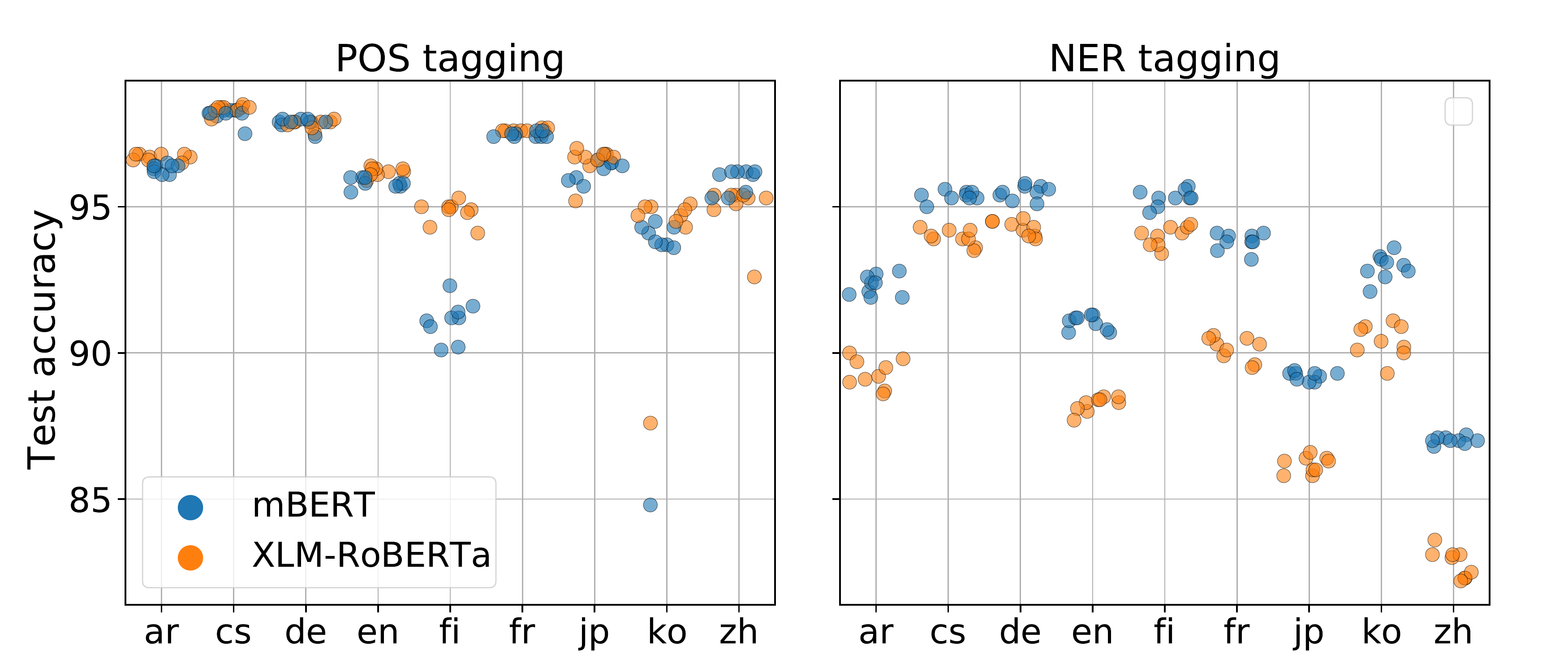}
  \end{center}
  \caption{\mbert vs.~\mroberta for POS tagging and NER.}
  \label{fig:swarm_pos_ner}
  \vspace*{-0.5cm}
\end{figure}

\paragraph{\mbert vs.~\mroberta} As with morphological probing tasks, \mroberta
is slightly better than \mbert (95.4 vs.~94.6 macro average). We also observe
that the choice of subword makes less difference than it does in morphological
probing.  Figure~\ref{fig:swarm_pos_ner} shows that experiments in one language
tend to cluster together regardless of the subword pooling choice except for a
few outliers: \swchoice{first} for Chinese and Korean is much worse in both
models.  The same result can be observed in Japanese, to a lesser extent
though. Language-wise we find that \mroberta is much better at Finnish and
somewhat worse in Chinese but the two models generally perform similarly.

\paragraph{Choice of subword.} As with morphology \swchoice{first} the is the
worst choice, but the effect is not as marked for POS tasks.  In
Figure~\ref{fig:swarm_pos_ner} we observe 3 outliers, \mroberta, \textsc{first}
for Chinese and \textsc{first} for Korean for both models. The only consistent
trend is that \mroberta is clearly better for Finnish regardless of the choice
of subword pooling. The picture is less clear for other languages.

\begin{figure*}[ht]
  \begin{center}
      \includegraphics[width=.99\textwidth,trim={0cm 0.8cm 0cm 0.3cm}]{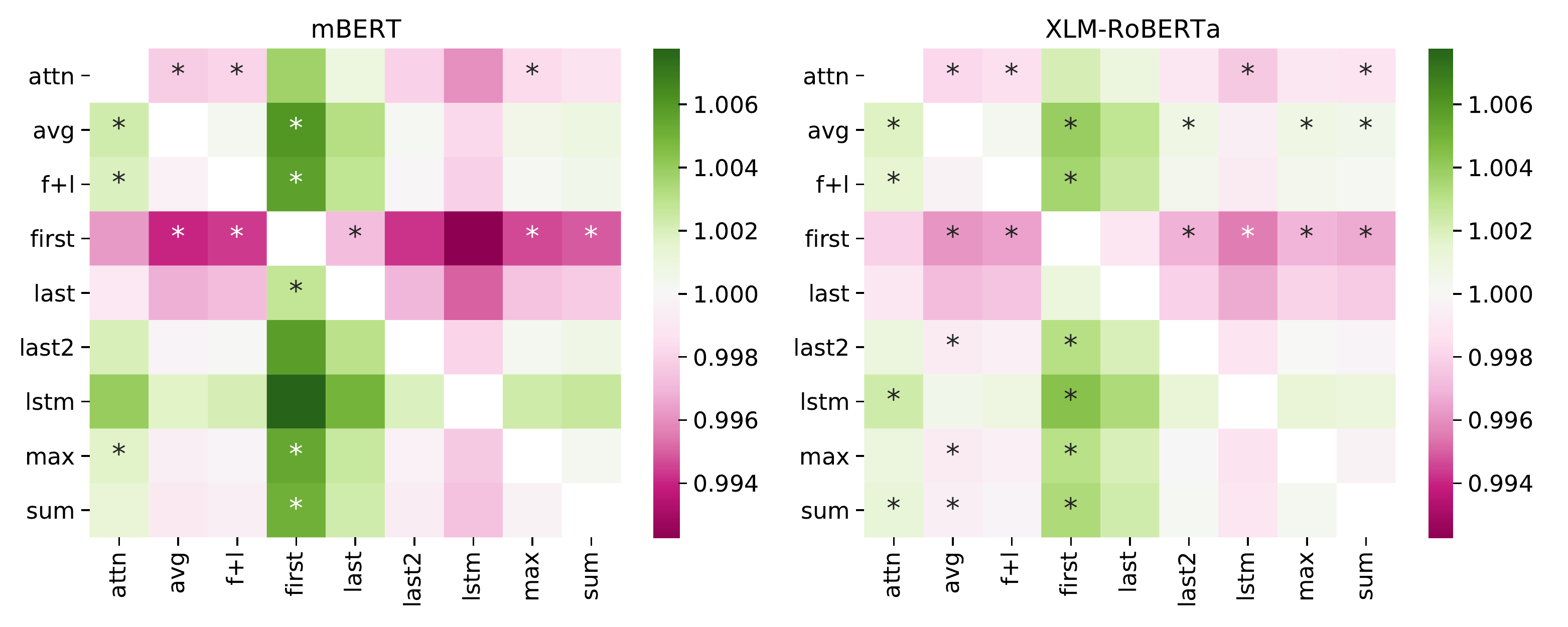}
  \end{center}
  \caption{Subword choices for POS tagging in non-CJK languages. See
    Figure~\ref{fig:heatmap_pair_morph} for an explanation of the figure.
    \swchoice{first} is included.}
  \label{fig:heatmap_pair_pos_no_cjk}
\end{figure*}

\begin{figure*}[ht]
  \begin{center}
      \includegraphics[width=.99\textwidth,trim={0cm 0.8cm 0cm 0.3cm}]{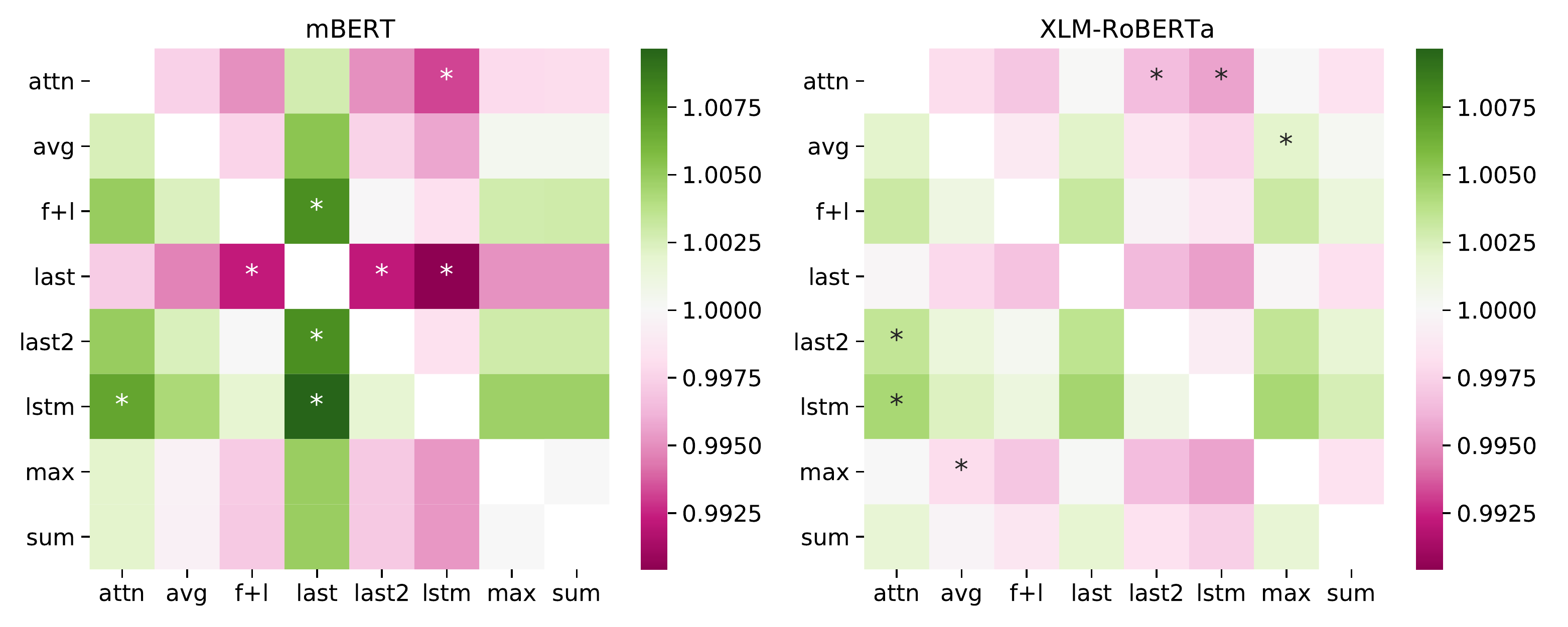}
  \end{center}
  \caption{Subword choices for POS tagging in CJK languages. See
    Figure~\ref{fig:heatmap_pair_morph} for an explanation of the figure.
    \swchoice{first} is omitted for clarity as it is much worse than the other
    choices.}
  \label{fig:heatmap_pair_pos_cjk}
\end{figure*}

We split the analysis into CJK and non-CJK languages.
Figure~\ref{fig:heatmap_pair_pos_no_cjk} and
Figure~\ref{fig:heatmap_pair_pos_cjk} show a comparison for non-CJK languages
and CJK languages respectively. The difference between choices is generally
much smaller than for morphology. \swchoice{first} is the worst choice both for
CJK and non-CJK languages.  Interestingly one of the best choices for morphology,
\swchoice{last}, is the second worst choice for POS tagging, while one of the worst
for morphology, \swchoice{lstm}, is one of the best for POS tagging. We
hypothesize that this is due to overparametrization for morphology. POS tagging
is a much more complex task that needs a larger number of trainable parameters
(recall that LSTM parameters are shared across all tokens).

%

\subsection{Named entity recognition}

As Figure~\ref{fig:swarm_pos_ner} shows, in NER the choice of subword pooling
makes far less difference than in morphology. In terms of models, \mbert has a
clear advantage over \mroberta when it comes to NER. The difference between
the two models is generally larger than the difference between two subword
choices within the same language. The smallest difference between the two
models appears to be in Czech, Finnish and German, which all have rich,
partially agglutinative, morphology. This fits with our earlier findings that
showed that \mroberta might be better at handling rich
morphology.
Overall \swchoice{first} and the related \swchoice{f+l} as well as
\swchoice{lstm} come out as winners, the differences are rather small and
often not statistically significant for CJK.

\subsection{Discussion}

Throughout our extensive experiments we observed that pooling strategies can
have a significant impact on the conclusions drawn from probing experiments.
When considering multiple typologically different languages, the strength of
the conclusions drawn from experiments can be weakened by considering a single
pooling option.  Our recommendation for NLP practitioners is to try at least
three subword pooling strategies, particularly for tasks in languages other
than English. \swchoice{first} and \swchoice{last} usually gives a general
picture -- as a third control we recommend \swchoice{attn} and
\swchoice{lstm}.  More complicated tasks such as POS or NER tagging may
require \swchoice{lstm} with many parameters, while tasks that rely more on
the orthographic representation such as morphology tend to benefit from
\swchoice{attn}.

One of the greatest attractions of the current generation of models is that
they do away with labor-intensive feature engineering. Currently, subword
pooling acts as the little finderscope mounted on the side of the main
telescope to get it to point in the right region, but over the long haul we
expect the systems to develop in a way that pooling also becomes part of the
end to end process.

Our methodology is only limited by the availability of data. It would be
interesting to extend these study with languages that use prefixes too such as
Indonesian or Swahili.

\section{Conclusion}\label{sec:conclusion}

The key takeaway from our work is that performance on lower level tasks
depends on the way we pool over multiple subword units that belong in a single
word token.  This is more of an
issue in languages other than English, where a significantly larger proportion
of words are represented by multiple subword units.

Morphological and POS tasks are both probing word-level attributes, but the
results show huge disparity: for the morphological tasks {\sc first} pooling is
the worst strategy, and {\sc attn} is the best, while for POS tagging {\sc
attn} is almost as bad as {\sc first}, the best being {\sc lstm}. The NER task
is intermediary between word- and phrase-level, and subword pooling effects are
less marked, but still statistically significant (see the full result tables in
the Appendix). 

\section*{Acknowledgments}

This work was partially supported by the BME Artificial
Intelligence TKP2020 IE grant of NKFIH Hungary (BME IE-MI-SC TKP2020) and by
the Hungarian Ministry of Innovation and the National Research, Development and
Innovation Office within the framework of the Artificial Intelligence National
Laboratory Programme.

\bibliography{ml}
\bibliographystyle{aclnatbib}

\appendix
\section*{Appendices}

\section{Data preparation}\label{app:data}

\subsection{Morphological probes}

We extract 2000 train, 200 dev and 200 test sentences for each task. We keep
UD's original splits, in other words, all of our train sentences come from UD's
train set.  We sample the sentences in a way that avoids overlaps in target
words between train, dev and test splits, in other words, if a word is the
target in the train set, we do not allow the same target word in the dev or
test set. A target word is the word that needs to be classified according to
some morphological tag. We also limit class imbalance to 3:1 at max. This
results in the removal of rare tags such as a few of the numerous Finnish noun
cases. These restrictions and the size of the treebanks do not allow generating
larger datasets.

\subsection{POS dataset}

We use the largest treebank in each language for POS. The only preprocessing we
do is that we filter sentences longer than 40 tokens. Since this results in an
uneven distribution in the training size, we limit the number of training
sentences to 2000. We note that experiments using 10,000 sentences are underway
but due to resource limitations, we were unable to include them in this version
of the paper.

\subsection{NER dataset}

NER is sampled from WikiAnn. WikiAnn is a silver standard large scale NER
corpus and the number of sentences is over than 100,000 in each language.
We deduplicated the dataset and discarded sentences longer than 40 tokens or
200 character in the case of Chinese and Japanese. WikiAnn annotates Chinese
and Japanese at the character level. We aligned this with mBERT's tokenizer and
retokenized it.  Due to memory constraints, we had to cut off the training data
size at 10,000.

\section{Training details}\label{app:training}

Each classifier is trained separately from randomly initialized
weights with the Adam optimizer \cite{Kingma:2014} with $(lr=0.001,
\beta_1=0.9, \beta_2=0.999)$ and early stopping on the development set.  We
report test accuracy scores averaged over 3 runs with different random seeds.

We ran about 14,000 experiments on GeForce RTX 2080 GPUs which took 7 GPU days.
We cache mBERT's and \mroberta's output when possible. We used PyTorch and our
own framework for experiment management. We release the framework along
with the final submission.

\section{Full result tables}\label{app:full_results}

\begin{table*}[!t]
    \small
    \centering
  \begin{tabular}{llllllllllr}
\toprule
                 \textbf{task} & \textbf{model} & \textsc{first} & \textsc{last} & \textsc{last2} & \textsc{f+l} & \textsc{sum} & \textsc{max} & \textsc{avg} & \textsc{attn} &  \textsc{lstm} \\
\midrule
      $\langle$Arabic, Case, NOUN$\rangle$ &          mBERT &     \textbf{76.5} &          73.3 &             74 &         74.8 &           75 &         73.5 &           75 &          75.5 &           74.8 \\
      $\langle$Arabic, Case, NOUN$\rangle$ &         XLM-Ro &           77.1 &          74.5 &             74 &         75.6 &   \textbf{80.9} &         80.1 &         79.3 &          77.8 &           74.5 \\
     $\langle$Arabic, Gender, ADJ$\rangle$ &          mBERT &           93.3 &          97.3 &             97 &         97.5 &         98.8 &         98.2 &         98.2 &    \textbf{99.3} &           97.3 \\
     $\langle$Arabic, Gender, ADJ$\rangle$ &         XLM-Ro &           96.5 &          97.5 &           97.2 &         97.5 &           99 &           99 &           99 &    \textbf{99.5} &           97.5 \\
      $\langle$Czech, Gender, ADJ$\rangle$ &          mBERT &           41.5 &    \textbf{77.1} &             74 &         72.6 &         64.2 &           63 &           63 &          75.6 &           68.0 \\
      $\langle$Czech, Gender, ADJ$\rangle$ &         XLM-Ro &           41.6 &    \textbf{73.5} &           71.8 &         70.8 &         64.2 &           65 &         63.2 &            73 &           66.5 \\
     $\langle$Czech, Gender, NOUN$\rangle$ &          mBERT &           61.7 &          77.4 &           77.9 &         74.6 &         76.3 &         75.1 &         74.8 &    \textbf{81.3} &           79.3 \\
     $\langle$Czech, Gender, NOUN$\rangle$ &         XLM-Ro &           69.3 &    \textbf{84.1} &           83.3 &         83.7 &         80.6 &         82.4 &         81.1 &          82.9 &           82.4 \\
 $\langle$English, Verbform, VERB$\rangle$ &          mBERT &           83.7 &    \textbf{91.2} &           87.2 &         89.2 &           89 &         89.5 &         89.7 &          90.7 &           88.5 \\
 $\langle$English, Verbform, VERB$\rangle$ &         XLM-Ro &           82.7 &            91 &           89.7 &         90.8 &   \textbf{91.8} &         91.7 &           89 &          91.3 &           89.7 \\
     $\langle$Finnish, Case, NOUN$\rangle$ &          mBERT &             33 &          90.7 &           90.7 &         87.7 &         87.4 &         86.2 &         88.1 &    \textbf{93.9} &           92.0 \\
     $\langle$Finnish, Case, NOUN$\rangle$ &         XLM-Ro &           49.6 &          94.4 &           94.7 &         94.2 &         94.9 &         95.7 &         95.7 &    \textbf{96.0} &           95.4 \\
 $\langle$Finnish, Verbform, VERB$\rangle$ &          mBERT &           74.6 &          92.5 &           91.4 &         92.5 &         89.9 &           90 &         90.9 &    \textbf{93.2} &           91.9 \\
 $\langle$Finnish, Verbform, VERB$\rangle$ &         XLM-Ro &           88.6 &    \textbf{97.8} &           97.5 &   \textbf{97.8} &         96.7 &         97.3 &         97.2 &    \textbf{97.8} &           96.5 \\
     $\langle$French, Gender, ADJ$\rangle$ &          mBERT &           87.7 &    \textbf{93.2} &           92.8 &   \textbf{93.2} &         92.7 &         92.7 &         92.3 &          92.2 &           92.5 \\
     $\langle$French, Gender, ADJ$\rangle$ &         XLM-Ro &           83.2 &            92 &           91.7 &         90.5 &         89.7 &         89.3 &         88.7 &    \textbf{92.5} &           90.3 \\
    $\langle$French, Gender, NOUN$\rangle$ &          mBERT &           88.3 &    \textbf{96.7} &           94.7 &         95.2 &         95.3 &         95.2 &           96 &          95.8 &           96.2 \\
    $\langle$French, Gender, NOUN$\rangle$ &         XLM-Ro &           93.3 &          93.5 &           94.8 &   \textbf{96.0} &         95.8 &         95.7 &         95.2 &          95.8 &           95.8 \\
  $\langle$French, Verbform, VERB$\rangle$ &          mBERT &           95.5 &   \textbf{100.0} &           99.7 &         99.5 &         99.7 &         99.3 &         99.5 &          99.8 &           99.5 \\
  $\langle$French, Verbform, VERB$\rangle$ &         XLM-Ro &           93.5 &          99.3 &           99.2 &         99.2 &   \textbf{99.5} &   \textbf{99.5} &         99.3 &          99.3 &           99.0 \\
      $\langle$German, Case, NOUN$\rangle$ &          mBERT &           63.2 &    \textbf{77.7} &             72 &         75.3 &           74 &         74.5 &         75.3 &            77 &           74.0 \\
      $\langle$German, Case, NOUN$\rangle$ &         XLM-Ro &           64.7 &          71.8 &           70.8 &   \textbf{77.3} &         74.2 &           75 &         74.2 &          71.3 &           73.7 \\
     $\langle$German, Gender, ADJ$\rangle$ &          mBERT &           58.9 &          77.9 &     \textbf{78.1} &         73.6 &         67.7 &           70 &         71.6 &          76.1 &           73.0 \\
     $\langle$German, Gender, ADJ$\rangle$ &         XLM-Ro &           54.7 &          74.6 &     \textbf{75.0} &         74.5 &         73.8 &         71.8 &         71.8 &            74 &           69.7 \\
    $\langle$German, Gender, NOUN$\rangle$ &          mBERT &           60.5 &          79.1 &           78.8 &         78.3 &         77.8 &         78.6 &         78.4 &    \textbf{79.8} &           76.8 \\
    $\langle$German, Gender, NOUN$\rangle$ &         XLM-Ro &           61.2 &          84.4 &           85.7 &         82.8 &         84.1 &         85.6 &         85.1 &    \textbf{88.9} &           86.9 \\
  $\langle$German, Verbform, VERB$\rangle$ &          mBERT &           89.2 &          94.5 &           94.5 &   \textbf{95.0} &         94.5 &   \textbf{95.0} &   \textbf{95.0} &    \textbf{95.0} &           94.2 \\
  $\langle$German, Verbform, VERB$\rangle$ &         XLM-Ro &           90.4 &          94.4 &           94.7 &         94.2 &         94.4 &         93.9 &         94.4 &    \textbf{95.0} &           94.7 \\
\bottomrule
\end{tabular}

    \caption{Full list of morphological probing results at the 6th layer.}
  \label{tab:all_morph_results}
\end{table*}

\begin{table*}[!t]
    \small
    \centering
  \begin{tabular}{llrllrrllrl}
\toprule
\textbf{language} & \textbf{model} &  \textsc{attn} & \textsc{avg} & \textsc{f+l} &  \textsc{first} &  \textsc{last} & \textsc{last2} & \textsc{lstm} &  \textsc{max} & \textsc{sum} \\
\midrule
           Arabic &          mBERT &           96.1 &         96.4 &         96.4 &            96.1 &           96.2 &           96.3 &    \textbf{96.5} &          96.4 &         96.4 \\
           Arabic &    XLM-RoBERTa &           96.5 &   \textbf{96.8} &   \textbf{96.8} &            96.6 &           96.6 &           96.7 &    \textbf{96.8} &          96.7 &   \textbf{96.8} \\
          Chinese &          mBERT &           95.5 &   \textbf{96.2} &   \textbf{96.2} &            95.3 &           95.3 &     \textbf{96.2} &    \textbf{96.2} &          96.1 &         96.1 \\
          Chinese &    XLM-RoBERTa &           94.9 &         95.3 &   \textbf{95.4} &            92.6 &           95.3 &     \textbf{95.4} &    \textbf{95.4} &          95.1 &   \textbf{95.4} \\
            Czech &          mBERT &           98.2 &         98.2 &   \textbf{98.3} &            97.5 &           98.2 &     \textbf{98.3} &    \textbf{98.3} &          98.2 &         98.1 \\
            Czech &    XLM-RoBERTa &           98.4 &         98.4 &         98.4 &            98.0 &           98.3 &           98.4 &    \textbf{98.5} &          98.3 &         98.4 \\
          English &          mBERT &           95.8 &   \textbf{96.0} &   \textbf{96.0} &            95.5 &           95.7 &           95.8 &    \textbf{96.0} &          95.8 &         95.7 \\
          English &    XLM-RoBERTa &           96.1 &         96.3 &         96.2 &            95.9 &           96.1 &           96.3 &    \textbf{96.4} &          96.3 &         96.2 \\
          Finnish &          mBERT &           90.9 &         91.4 &         91.1 &            90.1 &           90.2 &           91.6 &    \textbf{92.3} &          91.2 &         91.2 \\
          Finnish &    XLM-RoBERTa &           94.8 &           95 &           95 &            94.3 &           94.1 &           94.9 &    \textbf{95.3} &          94.9 &           95 \\
           French &          mBERT &           97.4 &   \textbf{97.6} &         97.5 &            97.4 &           97.4 &           97.4 &          97.4 &          97.5 &   \textbf{97.6} \\
           French &    XLM-RoBERTa &           97.6 &   \textbf{97.7} &   \textbf{97.7} &            97.5 &           97.6 &           97.6 &          97.6 &          97.6 &         97.6 \\
           German &          mBERT &           97.8 &         97.9 &   \textbf{98.0} &            97.4 &           97.9 &     \textbf{98.0} &    \textbf{98.0} &          97.9 &         97.9 \\
           German &    XLM-RoBERTa &           97.7 &         97.9 &         97.9 &            97.5 &           97.9 &           97.9 &    \textbf{98.0} &          97.9 &         97.8 \\
         Japanese &          mBERT &           96.0 &         96.5 &         96.6 &            95.7 &           95.9 &           96.5 &    \textbf{96.8} &          96.4 &         96.3 \\
         Japanese &    XLM-RoBERTa &           96.4 &         96.8 &         96.7 &            95.2 &           96.6 &           96.8 &    \textbf{97.0} &          96.7 &         96.7 \\
           Korean &          mBERT &           94.1 &         93.7 &         94.3 &            84.8 &           93.6 &           94.3 &    \textbf{94.5} &          93.7 &         93.8 \\
           Korean &    XLM-RoBERTa &           94.9 &         94.7 &           95 &            87.6 &           94.3 &             95 &    \textbf{95.1} &          94.5 &         94.7 \\
\bottomrule
\end{tabular}

    \caption{Full list of POS tagging results at the 6th layer.}
  \label{tab:all_pos_results}
\end{table*}

\begin{table*}[!t]
    \small
    \centering
  \begin{tabular}{llrlllrrlrl}
\toprule
\textbf{language} & \textbf{model} &  \textsc{attn} & \textsc{avg} & \textsc{f+l} & \textsc{first} &  \textsc{last} &  \textsc{last2} & \textsc{lstm} &  \textsc{max} & \textsc{sum} \\
\midrule
           Arabic &          mBERT &           92.0 &         92.6 &   \textbf{92.8} &           92.4 &           91.9 &            91.9 &          92.7 &          92.1 &         92.4 \\
           Arabic &    XLM-RoBERTa &           88.6 &         89.8 &         89.7 &           89.1 &           88.7 &            89.0 &    \textbf{90.0} &          89.2 &         89.5 \\
          Chinese &          mBERT &           87.0 &           87 &           87 &             87 &           86.9 &            87.1 &          86.8 &          87.1 &   \textbf{87.2} \\
          Chinese &    XLM-RoBERTa &           82.2 &         83.1 &         83.1 &           82.5 &           82.3 &            83.1 &    \textbf{83.6} &          82.3 &           83 \\
            Czech &          mBERT &           95.3 &         95.5 &   \textbf{95.6} &           95.3 &           95.0 &            95.3 &          95.5 &          95.4 &         95.4 \\
            Czech &    XLM-RoBERTa &           93.6 &         94.2 &   \textbf{94.3} &           93.9 &           93.5 &            93.9 &          94.2 &          93.9 &           94 \\
          English &          mBERT &           90.8 &         91.2 &   \textbf{91.3} &             91 &           90.7 &            90.7 &    \textbf{91.3} &          91.1 &         91.2 \\
          English &    XLM-RoBERTa &           87.7 &   \textbf{88.5} &   \textbf{88.5} &           88.3 &           88.0 &            88.1 &          88.4 &          88.3 &         88.4 \\
          Finnish &          mBERT &           95.3 &         95.5 &         95.6 &           95.3 &           94.8 &            95.0 &    \textbf{95.7} &          95.3 &         95.3 \\
          Finnish &    XLM-RoBERTa &           93.7 &   \textbf{94.4} &         94.3 &             94 &           93.4 &            93.7 &          94.3 &          94.1 &         94.1 \\
           French &          mBERT &           93.8 &   \textbf{94.1} &   \textbf{94.1} &           93.8 &           93.2 &            93.5 &            94 &          94.0 &         93.8 \\
           French &    XLM-RoBERTa &           89.9 &         90.5 &   \textbf{90.6} &           90.1 &           89.5 &            89.6 &          90.5 &          90.3 &         90.3 \\
           German &          mBERT &           95.4 &         95.7 &   \textbf{95.8} &           95.6 &           95.1 &            95.2 &          95.7 &          95.5 &         95.5 \\
           German &    XLM-RoBERTa &           94.0 &         94.5 &         94.5 &           94.3 &           93.9 &            94.0 &    \textbf{94.6} &          94.2 &         94.4 \\
         Japanese &          mBERT &           89.2 &         89.3 &         89.3 &     \textbf{89.4} &           89.0 &            89.0 &          89.3 &          89.3 &         89.1 \\
         Japanese &    XLM-RoBERTa &           85.8 &         86.4 &         86.4 &           86.3 &           85.8 &            86.0 &    \textbf{86.6} &          86.0 &         86.3 \\
           Korean &          mBERT &           92.8 &         93.3 &         93.2 &           92.8 &           92.1 &            92.6 &    \textbf{93.6} &          93.0 &         93.1 \\
           Korean &    XLM-RoBERTa &           90.2 &         90.9 &         90.9 &             90 &           89.3 &            90.1 &    \textbf{91.1} &          90.4 &         90.8 \\
\bottomrule
\end{tabular}

    \caption{Full list of NER results at the 6th layer.}
  \label{tab:all_ner_results}
\end{table*}

\end{document}